\documentclass[10pt]{article}

\newcommand{\ModelName}{Iris-Z}
\newcommand{\model}{\ModelName}
\newcommand{\modelbig}{\ModelName-big}
\newcommand{\modelsmall}{\ModelName-small}
\newcommand{\ReportTitle}{\ModelName: Scaling One Generative Model Past Specialist Geometry Models}
\newcommand{\ReportTopic}{Generalist models · Spatial intelligence}               
\newcommand{\ReportShort}{\MakeUppercase{\ModelName} TECHNICAL REPORT} 
\newcommand{\ReportYear}{2026}
\newcommand{\ReportCodeURL}{https://github.com/speridlabs/iris-z-sdk}
\newcommand{\ReportCodeLabel}{github.com/speridlabs/iris-z-sdk}
\newcommand{\ReportAuthors}{\textbf{Alejandro Ca\~nada}\textsuperscript{\textdagger}, \textbf{Chema Garabito}}
\newcommand{\ReportAuthorsPlain}{Alejandro Ca\~nada\textsuperscript{\textdagger}, Chema Garabito}

\usepackage[paper=a4paper,top=18mm,bottom=20mm,left=20mm,right=20mm]{geometry}
\usepackage{amsmath,amssymb}
\usepackage{etoolbox}
\usepackage{latexml}
\usepackage{fontspec}
\usepackage{graphicx}
\usepackage{booktabs}
\usepackage[table]{xcolor}
\usepackage{caption}
\usepackage{float}
\usepackage{titlesec}
\usepackage{fancyhdr}
\usepackage{enumitem}
\usepackage{multicol}
\usepackage{balance}
\usepackage{adjustbox}
\usepackage{ragged2e}
\usepackage[skip=0.55em plus 0.1em, indent=0pt]{parskip}
\usepackage[colorlinks=true, linkcolor=ink, citecolor=g500, urlcolor=ink]{hyperref}

\AtBeginDocument{\appto\normalsize{%
  \setlength{\abovedisplayskip}{10pt plus 2pt minus 2pt}%
  \setlength{\belowdisplayskip}{10pt plus 2pt minus 2pt}%
  \setlength{\abovedisplayshortskip}{8pt plus 2pt}%
  \setlength{\belowdisplayshortskip}{8pt plus 2pt minus 2pt}}}

\definecolor{ink}{HTML}{0A0A0A}
\definecolor{g50}{HTML}{F7F7F7}
\definecolor{g100}{HTML}{EEEEEE}
\definecolor{g200}{HTML}{DDDDDD}
\definecolor{g300}{HTML}{BEBEBE}
\definecolor{g400}{HTML}{909090}
\definecolor{g500}{HTML}{6B6B6B}
\definecolor{g600}{HTML}{4A4A4A}
\definecolor{bestcell}{HTML}{EEEEEE}
\definecolor{runnerupcell}{HTML}{F7F7F7}
\newcommand{\best}[1]{\cellcolor{bestcell}\textbf{#1}}

\arrayrulecolor{g200}

\newfontfamily\inter[Path=fonts/,BoldFont=Inter-SemiBold.ttf]{Inter-Regular.ttf}
\newfontfamily\intermed[Path=fonts/]{Inter-Medium.ttf}
\newfontfamily\brock[Path=fonts/,BoldFont=Brockmann-Bold.otf,ItalicFont=Brockmann-SemiBoldItalic.otf,LetterSpace=-2.5]{Brockmann-SemiBold.otf}
\providecommand{\inter}{}\providecommand{\intermed}{}\providecommand{\brock}{}
\providecommand{\addfontfeature}[1]{}
\newcommand{\eyebrow}[1]{{\intermed\fontsize{7.2}{9}\selectfont\addfontfeature{LetterSpace=14}\color{g500}\MakeUppercase{#1}}}
\newcommand{\eyebrowmono}[1]{{\ttfamily\fontsize{6.8}{9}\selectfont\addfontfeature{LetterSpace=14}\color{g500}\MakeUppercase{#1}}}

\newcommand{\twodigits}[1]{\ifnum#1<10 0\fi\number#1}

\renewcommand{\thesubsection}{\arabic{section}.\arabic{subsection}}
\iflatexml\else
\titleformat{\section}{\brock\fontsize{15.5}{18}\selectfont\color{ink}}{\twodigits{\value{section}}\;·\;}{0pt}{}
\titlespacing*{\section}{0pt}{1.9em plus 0.3em}{0.55em}
\titleformat{\subsection}{\inter\bfseries\fontsize{10.5}{13}\selectfont\color{ink}}{\thesubsection}{0.5em}{}
\titlespacing*{\subsection}{0pt}{1.3em plus 0.2em}{0.3em}
\titleformat{\paragraph}[runin]{\inter\bfseries\color{ink}}{}{0pt}{}
\titlespacing*{\paragraph}{0pt}{0.7em}{0.45em}
\fi

\iflatexml\else
\DeclareCaptionLabelFormat{sl}{{\eyebrowmono{#1 #2}}}
\DeclareCaptionFormat{slfig}{\noindent{\color{g200}\rule{\linewidth}{0.4pt}}\par\vspace{4pt}\noindent#1\hspace{0.7em}#3\par}
\DeclareCaptionFormat{sltab}{\noindent#1\hspace{0.7em}#3\par\vspace{3pt}}
\fi

\newcommand{\thd}[1]{{\ttfamily\fontsize{6.8}{9}\selectfont\addfontfeature{LetterSpace=14}\color{g500}\MakeUppercase{#1}}}
\renewcommand{\toprule}{}
\renewcommand{\midrule}{\specialrule{1.4pt}{2pt}{4pt}\arrayrulecolor{g200}}
\renewcommand{\bottomrule}{\specialrule{0.4pt}{2pt}{0pt}}
\renewcommand{\arraystretch}{1.35}
\AtBeginEnvironment{tabular}{\small\arrayrulecolor{ink}}

\fancypagestyle{first}{\fancyhf{}\fancyfoot[L]{\eyebrowmono{Preprint · \ReportYear{} · Speridlabs Research}}\fancyfoot[R]{\eyebrowmono{\ReportShort{} · \thepage}}}
\setlist[itemize]{leftmargin=1.2em,itemsep=0.3em,label={\color{g400}\rule[0.5ex]{0.5em}{0.5pt}}}

\newsavebox{\abstractbox}
\newlength{\abstractH}
\iflatexml\else
\renewenvironment{abstract}{%
  \begin{lrbox}{\abstractbox}\begin{minipage}{\dimexpr\linewidth-14.2mm\relax}%
  \vspace{4mm}\setlength{\parskip}{0.5em}\small\linespread{1.22}\selectfont%
  \noindent\eyebrow{Abstract}\par\vspace{0.6em}}%
  {\par\vspace{4mm}\end{minipage}\end{lrbox}%
  \setlength{\abstractH}{\dimexpr\ht\abstractbox+\dp\abstractbox\relax}%
  \noindent{\setlength{\fboxsep}{0pt}\setlength{\fboxrule}{0.4pt}%
  \fcolorbox{g200}{g50}{\raisebox{-\dp\abstractbox}{\color{ink}\rule{2.2pt}{\abstractH}}\hspace{6mm}\usebox{\abstractbox}\hspace{6mm}}}\par\vspace{6mm}}
\fi

\newcommand{\preprinttitle}{%
  \setlength{\parskip}{0pt}%
  \noindent\begin{minipage}[b]{0.5\linewidth}\raisebox{-0.35\height}{\includegraphics[height=6.2mm]{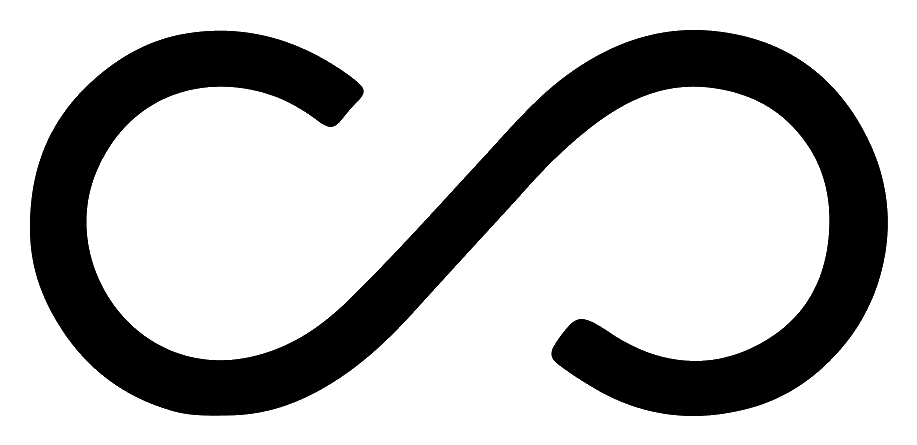}}\hspace{2.2mm}{\inter\fontsize{12}{14}\selectfont\color{ink}speridlabs}\end{minipage}%
  \begin{minipage}[b]{0.5\linewidth}\raggedleft\eyebrowmono{Preprint · Technical report}\\[2pt]\eyebrowmono{\ReportYear{} · Speridlabs Research}\end{minipage}\par
  \vspace{3mm}\noindent{\color{ink}\rule{\linewidth}{1.4pt}}\par\vspace{7mm}%
  \noindent\eyebrow{\ReportTopic}\par\vspace{2.5mm}%
  \begin{flushleft}\brock\fontsize{25}{27}\selectfont\color{ink}\ReportTitle\end{flushleft}\vspace{2.5mm}%
  \begin{flushleft}\inter\fontsize{10.5}{14}\selectfont\color{ink}\ReportAuthors\\[0.5em]
  {\footnotesize\color{g500}SperidLabs\;·\;\textsuperscript{\textdagger}Project lead\;·\;Code and models: \href{\ReportCodeURL}{\color{ink}\ReportCodeLabel}}\end{flushleft}\vspace{2mm}%
  \setlength{\parskip}{0.55em plus 0.1em}}

\makeatletter
\iflatexml\else
\renewcommand\@biblabel[1]{{\ttfamily\fontsize{7}{9}\selectfont\color{g500}[\ifnum#1<10 0\fi#1]}}
\renewenvironment{thebibliography}[1]{%
  \section*{References}\addcontentsline{toc}{section}{References}%
  \fontsize{8}{10.5}\selectfont\color{g600}%
  \list{\@biblabel{\@arabic\c@enumiv}}{\settowidth\labelwidth{\@biblabel{#1}}%
  \leftmargin\labelwidth\advance\leftmargin\labelsep\itemsep 0.1em\parsep 0pt\usecounter{enumiv}}%
  \sloppy\clubpenalty4000\widowpenalty4000}{\endlist}
\fi
\makeatother

\definecolor{pendingred}{HTML}{C0392B}

\title{\ReportTitle}
\author{\ReportAuthorsPlain\\
SperidLabs \quad \textsuperscript{\textdagger}Project lead \quad Code and models:
\href{\ReportCodeURL}{\ReportCodeLabel}}
\date{}
\iflatexml
\renewcommand{\thd}[1]{#1}\renewcommand{\eyebrowmono}[1]{#1}
\renewcommand{\eyebrow}[1]{#1}
\fi

\begin{document}
\iflatexml\maketitle\else\preprinttitle\thispagestyle{first}\fi

\begin{abstract}
We present \model{}, a single generative model that produces images and their depth together. The same set of weights
generates an image and its depth from a prompt, estimates depth from a photograph and paints an image onto a given
depth map. Monocular depth estimation has long been the domain of specialised architectures, and we ask whether a
general generative model can match them without any task-specific design. \model{} adds no specialised layers and is
trained with one generative objective. With enough scale it surpasses state-of-the-art specialist models on
established depth benchmarks while keeping the image-generation abilities of the backbone it started from. These
results suggest that, as in language, effort in visual and spatial understanding is better spent on general models
trained at scale than on task-specific architectures. We release two models to the community.
\end{abstract}

{\centering\includegraphics[width=\linewidth,height=\dimexpr\pagegoal-\pagetotal-15mm\relax,keepaspectratio,alt={Top: images generated from prompts, each above the depth map generated with it. Bottom: photographs above the depth the model estimates for them.}]{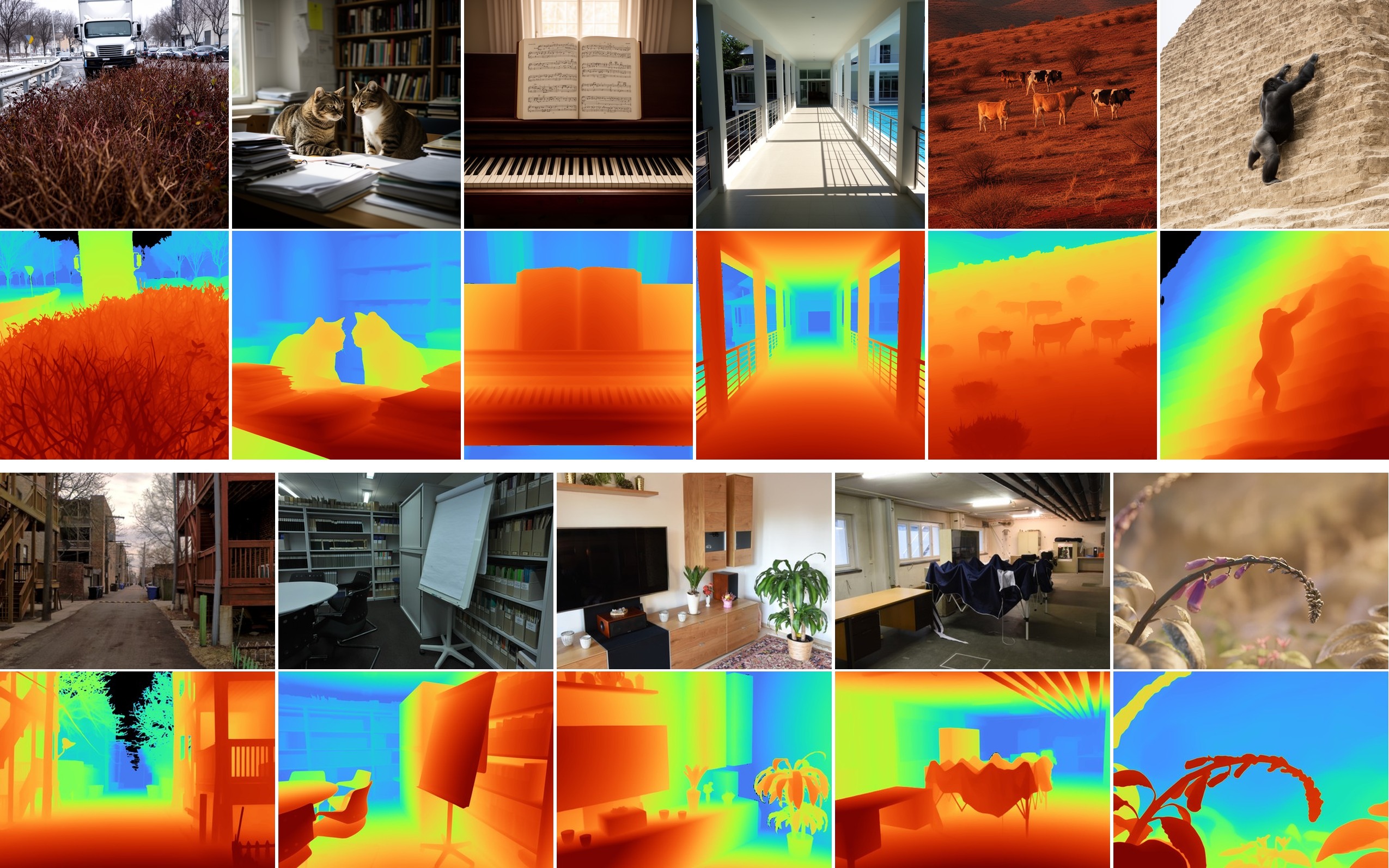}\par}\vspace{1mm}
\noindent{\footnotesize\color{g500}\eyebrowmono{Fig. 1}\hspace{0.7em}Top: prompt
$\rightarrow$ image and depth, generated together. Bottom: photographs $\rightarrow$ depth. All from \modelbig{}. Depth is colored on log
distance, near = red, far = blue; black: no depth or sky. }\par\vspace{5mm}
\setcounter{figure}{1}

\begin{multicols}{2}

\section{Introduction}
\label{sec:intro}
Language moved from many task-specific models to one model trained on a single generative task. Trained to
predict the next token on enough data, it learned to translate, summarize and answer questions about as well as
models built for each of those tasks~\cite{gpt3}, and it kept improving on all of them as data, parameters and
compute grew~\cite{kaplan2020,chinchilla}. This is the bitter lesson~\cite{bitterlesson}. General methods that
scale with computation end up ahead of task-specific ones.

We expect vision to follow the same path, from 2D perception to 3D and spatial understanding, and visual
generative models already show it. For example, Vision Banana instruction-tunes the Nano Banana Pro image generator on a small
amount of task data and matches or beats specialists such as Segment Anything~3 and Depth Anything without losing
its ability to generate~\cite{visionbanana}. Veo~3 segments objects, finds edges and solves simple visual tasks
without being trained for any of them~\cite{wiedemer2025}. BAGEL~\cite{bagel} and PixelUMM~\cite{pixelumm} go
further and use the same weights both to understand and to generate images.

Data is the other half of the argument. Geometry labels are scarce and expensive to collect, since they come from
depth sensors, laser scans or synthetic renders, while a generative model learns from web-scale images and videos
that carry no labels made for one task. In a unified architecture such as \model{}, where each modality is one more
group of tokens in the same sequence, new modalities can join that training and build on what the model already
learned from that data. In our experiments, a larger pre-trained generator trained on more data scored better on
depth benchmarks (Fig.~\ref{fig:glance}b).

Monocular geometry is the right place to test this hypothesis because generative models have already been applied
to it, and so far the specialists stay ahead. Image generators repurposed as depth estimators, such as the Marigold
family~\cite{marigold,marigoldv2}, although strong, still perform worse than both MoGe-3~\cite{moge3} and MoGe-2 on
MoGe-3's held-out test sets. MoGe-3, a ViT designed for this task, leads the standard benchmarks, ahead of Depth
Pro~\cite{depthpro}, Depth Anything~3~\cite{da3}, UniDepth~V2 and UniK3D in its own comparison. Vision Banana
reports a lower metric-depth error than MoGe-2~\cite{moge2}, but it predates MoGe-3 and is not available for
testing.

We built \model{} to test this hypothesis directly. Starting from a pre-trained image generator, we trained it to
generate and reason about geometry (depth) as well as appearance (RGB). In \model{}, appearance and geometry are
produced together as one RGBD output, denoised by a single backbone. Our view is that a generative model should
generate and understand the world in all its modalities jointly, and that the geometry it generates with an image is
part of what it knows about the scene.

We see this as a step toward a single model for visual understanding. The same recipe extends to other modalities,
such as surface normals, segmentation, motion or camera parameters, without changing the architecture, and each one
added becomes part of what the model learns about the scene.

\modelbig{} is, to our knowledge, the first open general-purpose generator to beat MoGe-3 on Marigold's benchmarks.
It is also ahead of MoGe-3 on average on MoGe-3's own benchmarks at the resolutions it was trained for, and it still
performs the generation tasks it handled before our training. Each step of scale we added (more data, a full
fine-tune, a larger backbone) lowered the error without any depth-specific layer (Fig.~\ref{fig:glance}b). We read
this as evidence that general generative models can overtake specialized geometry models as they scale.

Our contributions are:
\begin{itemize}
\item \model{}, a generative model that denoises image and depth jointly in one sequence and serves four
tasks with one set of weights, released in two sizes, \modelbig{} and \modelsmall{}.
\item State-of-the-art affine-invariant depth performance on Marigold's and MoGe-3's benchmarks (Sec.~\ref{sec:results}).
\item Evidence that the improvement in depth estimation comes from scale rather than task design. Across three runs we changed only the training data and the backbone size, and each run lowers the depth error (Sec.~\ref{sec:results}).

\end{itemize}

\end{multicols}
\begin{center}
\includegraphics[width=\linewidth,alt={Three panels. a: mean AbsRel on Marigold's benchmarks for Marigold V2 (4.10), MoGe-3 (3.12), Iris-Z-small (3.11) and Iris-Z-big (2.92). b: the same benchmark as the approach is scaled from a LoRA that saw 0.2M frames to full fine-tunes that saw 3 to 4M unique frames and a larger backbone. c: accuracy against DiT passes per image.}]{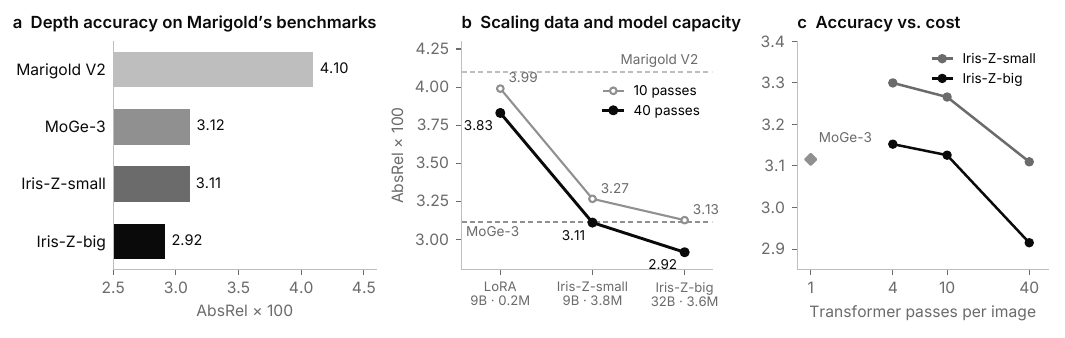}
\captionof{figure}{\model{} on Marigold V2's benchmarks, mean AbsRel$\times$100 (lower is better). \textbf{a}: Our two models vs Marigold V2 and MoGe-3. \textbf{b}: scaling: a LoRA on \modelsmall{}'s backbone that saw 0.2M
frames, a full fine-tune of it (\modelsmall{}) that saw 3.8M unique frames, then \modelbig{}, a backbone
3.5$\times$ larger, that saw 3.6M. Black: 40 inference passes, the sampler we report; gray: 10 passes. \textbf{c}: accuracy against the number of DiT passes per image (4: one seed,
10: 2 steps from 5 seeds, 40: 4 steps from 10 seeds). \modelbig{}'s 10-pass points are from an earlier checkpoint of the same training.}
\label{fig:glance}
\end{center}
\begin{multicols}{2}

\section{Related Work}
\label{sec:related}
\paragraph{Monocular depth specialists.} Learning depth from a single image began with networks regressed on one
dataset at a time, with a scale-invariant loss to deal with the ambiguity of a single view~\cite{eigen2014}. MiDaS
showed that mixing many datasets with different scales trained with
a loss invariant to scale and shift transferred to unseen domains far better than any single
source~\cite{midas}. DPT moved the same recipe to a vision transformer~\cite{dpt}. Metric depth needed camera-aware approaches like ZoeDepth which adds metric heads to a relative model~\cite{zoedepth}, Metric3D that maps
every image to a canonical camera~\cite{metric3d} or UniDepth that predicts the camera together with the
depth~\cite{unidepth}. Depth Anything scaled the training data instead, pseudo-labeling about 62M unlabeled images with a teacher~\cite{dav1}; its second version trains that teacher on synthetic data only~\cite{dav2}. Depth Pro then focused on sharp boundaries and focal length estimation~\cite{depthpro}, and Depth Anything~3 extended the approach to multiple views~\cite{da3}. MoGe predicts an affine-invariant point map with a robust alignment in
training~\cite{moge}, MoGe-2 adds metric scale~\cite{moge2}, and MoGe-3 refines the point map in 3D with
sparse convolutions~\cite{moge3}. This line of work improved by finding better depth-specific strategies and more varied and
better labeled data, while each model stayed task-specific.

\paragraph{Generative priors for dense prediction.} Marigold showed that a text-to-image latent diffusion model fine-tuned on a small
amount of synthetic data becomes a strong zero-shot depth estimator~\cite{marigold}. Much of the follow-up work made this faster:
DepthFM replaces diffusion with flow matching from image to depth~\cite{depthfm}, Lotus predicts the clean target in a single
step~\cite{lotus}, and Martin Garcia et~al.\ show that a single-step model 
matches the multi-step one~\cite{e2eft}. Other work extended it to new outputs and backbones. GeoWizard predicts depth and
normals together~\cite{geowizard}, DepthCrafter applies the approach to a video generator~\cite{depthcrafter}, Marigold~V2
moves to diffusion transformers~\cite{marigoldv2}, and Iris-3B compares pixel and latent-space generators~\cite{iris}. 
In all of these the generator becomes an image-to-depth model. The image is always the input, and the ability to 
generate images is lost along the way. \model{} keeps the generative objective and text-to-image generation, and 
depth estimation is one of several tasks the same joint model performs.
\paragraph{Joint RGB-D and multi-modal generation.} A smaller line of work produces geometry together with appearance. LDM3D trains one latent diffusion model to output an image and its depth from a prompt~\cite{ldm3d}, while JointNet keeps the text-to-image model frozen and adds a parallel branch for a dense modality such as depth~\cite{jointnet}. Wonder3D extends this across views, producing colors and normals with cross-domain attention~\cite{wonder3d}, and Aether trains a video model to predict geometry alongside the frames~\cite{aether}. None of these were shown to reach specialist accuracy on established depth benchmarks, which is the gap \model{} addresses.

\paragraph{Generalist vision through generation.} Visual in-context learning shows a model an example input and its output side by side together with a new input, and asks it to fill in the missing output, so the task is defined by the example rather than by the architecture. Bar et~al.\ do this with an inpainting model on a grid of images~\cite{visualprompting}, and Painter trains a masked image model on many tasks in this format~\cite{painter}. InstructCV instruction-tunes a text-to-image model to perform vision tasks named in text~\cite{instructcv}, and OmniGen trains one generator for editing, conditional generation and several classic vision tasks~\cite{omnigen}. More recently, Vision Banana showed that a frontier image generator with instruction-tuning matches specialists on segmentation and depth~\cite{visionbanana}, and Veo~3 solves visual tasks it was never trained for~\cite{wiedemer2025}. Unified multimodal models go further and both read and write images in one network. Transfusion combines next-token prediction and diffusion in one transformer~\cite{transfusion}, Janus uses separate visual encoders for understanding and generation~\cite{janus}, and BAGEL and PixelUMM train such models at larger scale~\cite{bagel,pixelumm}.

\paragraph{Scaling visual generators.} Diffusion transformers improve predictably with compute~\cite{dit}, and
rectified-flow transformers show smooth scaling of validation loss that tracks human preference up to billions of
parameters~\cite{sd3}. Whether the geometry a generator learns scales the same way is less studied. Our scaling path (Fig.~\ref{fig:glance}b) is a first data point for the depth modality.

\section{Method}
\label{sec:approach}
\model{} starts from a pretrained text-to-image diffusion transformer, and the only parameters we add are a learned
modality embedding, with no depth head and no second network. Depth is encoded as an image by the same autoencoder that encodes the image and placed in the 
same token sequence, and training uses the base model's flow-matching loss together with a loss on the decoded depth.
Below we describe how image and depth share
one sequence (Fig.~\ref{fig:method}), how depth is represented, which tasks are trained, and how we sample. The backbones, training settings
and data are in Sec.~\ref{sec:impl}.

\paragraph{Image and depth in one sequence.} The image and depth latents form one token sequence together with
the prompt. Each token carries three embeddings that identify it. The \emph{spatial position embedding} (the
backbone's rotary row and column) is the same for the image token and the depth token of a patch, so the model can
match tokens that describe the same point of the scene. The \emph{sequence position embedding}, an extra rotary
coordinate the backbone already uses to tell several input images apart, is 0 for image tokens and 1 for depth
tokens. A learned \emph{modality embedding} labels each token as generated image, generated depth, given image or
given depth. In addition, image and depth each carry their own \emph{flow-matching time}~$t$. Whichever is generated
follows the rectified-flow objective of the base model~\cite{flowmatching,sd3} at its own~$t$, and whichever is
given is fed clean and modulated with $t = 0$. Because attention runs over the whole sequence, image and depth
condition each other at every step of denoising.

\paragraph{Depth as an image.} Depth is stored as per-image normalized log ray distance. The 2nd to 98th
percentile of an image's log distance is mapped to $[-1, 0.95]$, so every image uses the same range and the
prediction is defined up to a scale and shift in log space. Sky is set to $+1$, which leaves a margin between the
farthest real depth and the sky. The normalized map is copied into three channels and encoded by the unchanged image
autoencoder, and decoding averages the three channels. At inference, pixels above $0.975$, halfway between the depth
range and the sky value, are marked as sky. These values are for \modelbig{}. \modelsmall{} uses the full
range $[-1, 1]$, with no margin reserved for sky.

\paragraph{Multiple tasks with the same weights.} Each training sample draws one of five tasks. In the
\emph{joint} task, image and depth are noised to the same time and both are scored (prompt $\rightarrow$ RGBD). With
the \emph{image given}, only the depth is denoised (image $\rightarrow$ depth), and with the \emph{depth given}, only
the image is (depth $\rightarrow$ image). In the \emph{independent} task, image and depth are noised to independent
times. In the \emph{image-only} task, the depth stays pure noise and only the image is scored (prompt $\rightarrow$
image), which keeps plain text-to-image generation in training and prevents the model from drifting away from the
base model's images. A second loss decodes the depth the model currently predicts and compares it with the target in
pixel space. Captions are dropped for a share of samples, so image $\rightarrow$ depth works without a prompt. With
this mix, one set of weights performs prompt $\rightarrow$ RGBD, image $\rightarrow$ depth, depth $\rightarrow$ image
and prompt $\rightarrow$ image at inference.

\end{multicols}
\begin{center}
\includegraphics[width=\linewidth,alt={Prompt, image and depth are encoded into one token sequence and processed together by one DiT backbone; the image and the depth go through the same VAE encoder, and the generated image and depth through the same VAE decoder. The pictures are a real generated image and its depth.}]{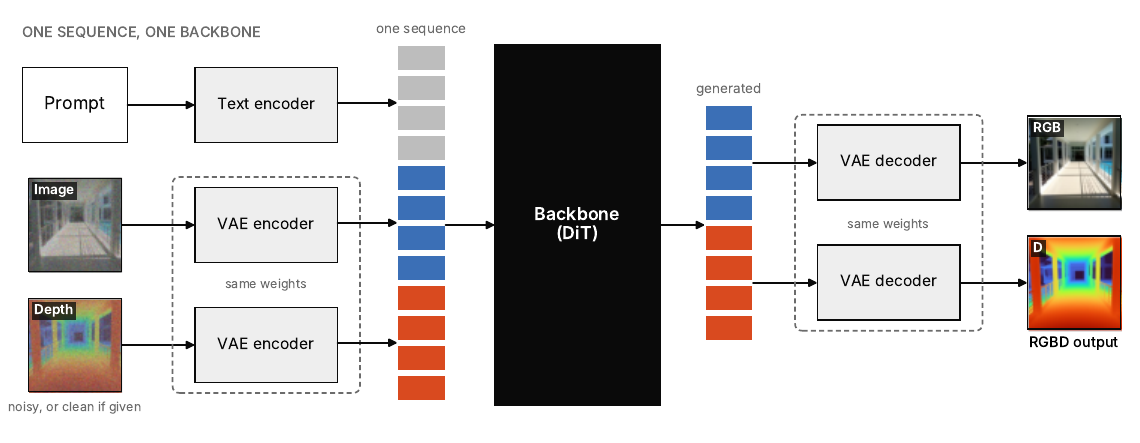}
\captionof{figure}{\model{}. The prompt, the image and the depth are encoded into one token sequence and processed
by a single DiT. Image and depth share the VAE encoder and decoder, and each enters as a noisy latent, or clean when the task provides it.}
\label{fig:method}
\end{center}
\begin{multicols}{2}

\paragraph{Sampling depth.} Image $\rightarrow$ depth denoises the depth with the image given and an empty prompt,
so both branches of classifier-free guidance are identical and each step costs one transformer pass. Because sampling
starts from noise, we combine a few short trajectories from different seeds (Sec.~\ref{sec:impl-sampling}).

\section{Implementation Details}
\label{sec:impl}

\subsection{Backbones}
\label{sec:impl-backbones}
We use two open text-to-image transformers from the FLUX.2 family as backbones~\cite{flux2dev,flux2klein}.
\modelbig{} starts from FLUX.2 [dev] (32B parameters, Mistral Small 24B text encoder) and \modelsmall{} from
FLUX.2 [klein] Base 9B (Qwen3-8B text encoder). Both are full fine-tunes of the transformer, with the autoencoder and
the text encoder frozen. FLUX.2 [dev] is guidance-distilled, so we remove its guidance embedding and train
classifier-free guidance back through caption dropout. The modality embedding is initialized at zero, so training
starts from the unchanged base model.

\subsection{Training}
\label{sec:impl-training}
Both models are trained with AdamW in bf16 mixed precision and with sharded data parallelism. The decoded-depth
loss takes the clean depth latent implied by the prediction, $\hat{x}_0 = x_t - t\,v$, decodes it with the frozen
autoencoder and computes an L1 to the target over valid pixels, weighted by $(1 - t)$ so that near-clean steps count
most. It is added with weight 1 to the flow-matching loss of every sample whose depth is scored. On sources with
sparse or holed depth (LiDAR, stereo, relative reconstructions), depth is supervised only by this decoded loss, on
measured pixels only. We keep two exponential moving averages of the weights. The slower one is better on every
benchmark for both models, and it is the one we release.

\modelbig{} trains at a constant learning rate that is annealed linearly to zero at the end of training, while
\modelsmall{} keeps a constant learning rate throughout. Both start at about $512^2$ pixels and move a growing share of samples
to about $1024^2$, ending at 42\% (\modelbig{}) and 56\% (\modelsmall{}).
\subsection{Data}
\label{sec:data}
\modelsmall{} was trained on about 3.8M unique RGB-D frames and \modelbig{} on about 3.6M, drawn from real indoor
captures, driving scenes with LiDAR and stereo, multi-view reconstructions, rendered scenes and generated images.
Every source was converted to ray distance, in meters where the source has a metric scale. Sensor holes and sparse
LiDAR are kept as holes rather than filled. Sky follows one rule per source and is stored as a fixed far value.
Every frame has a dense caption, used as the prompt, and images are stored at about $512^2$ and $1024^2$ pixels.
Validation frames are held out per source and split by scene, and we do not train on the benchmark datasets.

\subsection{Sampling and generation}
\label{sec:impl-sampling}
Image $\rightarrow$ depth runs 4 steps from each of 10 seeds. Each seed's prediction is the average of the
clean-depth estimates $\hat{x}_0 = x_t - t\,v$ from its last two steps. The seeds are aligned to a reference by
least-squares scale and shift in log space and reduced by the per-pixel median, for 40 transformer passes per image.
Images under 0.5\,MP run at about $512^2$ pixels, larger ones at about $1024^2$ (with one exception on MoGe-3's
benchmarks, Sec.~\ref{sec:eval}). The benchmark then fits its own
alignment to the result, as for any model.

Prompt $\rightarrow$ image and prompt $\rightarrow$ RGBD use 50 steps with classifier-free guidance (3.5 for
\modelbig{}, 4.0 for \modelsmall{}). For prompt $\rightarrow$ RGBD the depth of the joint generation can be used as
is, or re-estimated from the finished image with the image $\rightarrow$ depth sampler, which is sharper.

Depth $\rightarrow$ image denoises the image with the depth given clean, starting from Gaussian noise, with 50 steps
from one seed and classifier-free guidance between the caption and the empty prompt. Without a prompt there is no guidance and each step is one transformer pass.
Holes in the given depth are filled with the nearest measured value.

\subsection{Generation is preserved}
\label{sec:generation}
A depth estimator that forgets how to draw would be just another specialist. We generate with fixed held-out prompts and
seeds and compare with the untouched base model on the same inputs (Tab.~\ref{tab:gen}, Fig.~\ref{fig:gen}) to ensure that \model{} is
in fact a general model.

Prompt adherence holds as CLIP stays within 0.4 of the base model, slightly above it for \modelsmall{}. Images keep the
base model's subject and composition (similarity 0.88--0.89). Preference scores drop (HPSv2.1 24.3 against 26.7 for
\modelbig{}) as the models drift a little toward the look of their training data. That same drift makes \modelbig{}'s images more photographic (0.48 against 0.38).

\vspace{2mm}\noindent
\begin{minipage}[t]{\linewidth}\centering
\captionof{table}{Prompt $\rightarrow$ image against the untouched base model, same prompts and seeds, 50 steps with
guidance. CLIP~\cite{clip}: 100$\times$ the image--prompt cosine. Sim.: CLIP cosine to the base model's image
(1 = identical). Photoreal: CLIP's zero-shot probability that the image is a photograph rather than a render or
drawing. HPSv2.1~\cite{hps}: a learned human-preference score.}
\label{tab:gen}
\setlength{\tabcolsep}{6pt}
\begin{tabular}{@{}lrrrr@{}}
\toprule
\thd{Model} & \thd{CLIP} & \thd{Sim.} & \thd{Photoreal} & \thd{HPSv2.1}\\
\midrule
FLUX.2 [klein] & 27.46 & -- & 0.41 & 24.50\\
\modelsmall{} & 27.74 & 0.875 & 0.34 & 24.01\\
\midrule
FLUX.2 [dev] & 27.85 & -- & 0.38 & 26.71\\
\modelbig{} & 27.42 & 0.888 & 0.48 & 24.30\\
\bottomrule
\end{tabular}
\end{minipage}

\begin{center}
\includegraphics[width=\linewidth,alt={Prompts; top row: images from the untouched base model; bottom row: images from our model with the same prompt and seed.}]{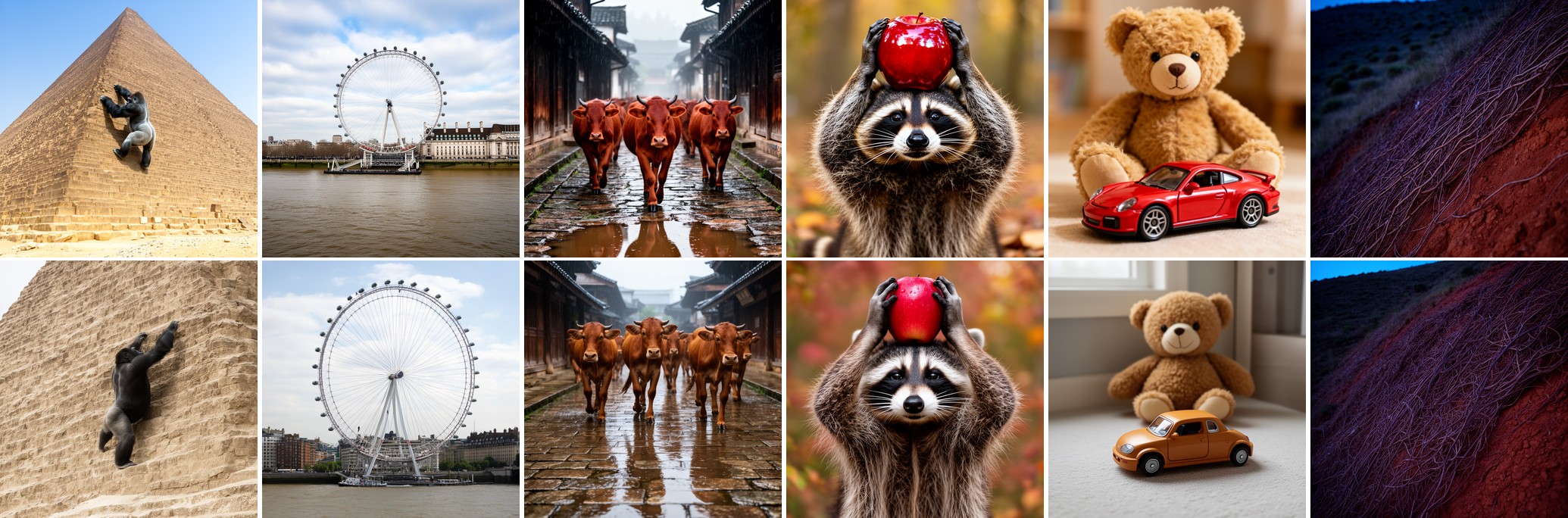}
\captionof{figure}{Same prompt and seed. Top: the untouched base model. Bottom: \model{}, the image from a joint
RGBD generation.}
\label{fig:gen}
\end{center}

\section{Evaluation}
\label{sec:eval}
\paragraph{Benchmarks.} We evaluate on two benchmark suites, each run exactly as its paper defines it. Marigold's
uses NYUv2, KITTI, ETH3D, ScanNet and DIODE with Marigold~V2's split lists and masks (3,331 images), scored after
RANSAC scale-and-shift alignment in log depth. MoGe-3's uses NYUv2, KITTI, ETH3D, iBims-1, Sintel, DIODE and HAMMER
with MoGe's masks (4,470 images), scored after MoGe's relative-error L1 alignment. We report AbsRel$\times$100.

\paragraph{Sets and resolution.} We leave out two of MoGe-3's nine sets, Spring (2\,MP) and Synth4K (8\,MP), because
both models were trained at up to about $1024^2$ pixels and these sets test resolutions we did not train for. ETH3D,
also high resolution, stays because it is one of Marigold~V2's benchmarks too. On Marigold's suite we keep the rule
of Sec.~\ref{sec:impl-sampling} (under 0.5\,MP at about $512^2$, larger at about $1024^2$). On MoGe-3's suite,
KITTI, whose ground truth is sparse LiDAR, also runs at about $1024^2$.

\paragraph{Baselines.} We run Marigold~V2 (a single-step model by design, one pass) and MoGe-3 (ViT-G, resolution
level 9, 3 refinement steps) with our code on the same pixels. Each table also lists the numbers published in
that benchmark's paper, for the models it reports.

\section{Results}
\label{sec:results}
\paragraph{MoGe-3's protocol.} On MoGe-3's own benchmarks and alignment (Tab.~\ref{tab:moge}, Fig.~\ref{fig:sets}),
\modelbig{} is the most accurate model on average (3.56 against 3.70 for MoGe-3) and on six of the seven sets, while
MoGe-3 keeps DIODE. \modelsmall{} is ahead of MoGe-3 on NYUv2 and HAMMER and behind on average (3.84).

\paragraph{Marigold's protocol.} \modelbig{} is the most accurate model on Marigold's benchmarks
(Tab.~\ref{tab:marigold}), at 2.92 against 3.12 for MoGe-3 and 4.10 for Marigold~V2. It is best on NYUv2, ETH3D and
ScanNet, while MoGe-3 keeps KITTI and DIODE. \modelsmall{}, at 9B, is level with MoGe-3 (3.11).

Scores on the datasets that appear in both protocols (NYUv2, KITTI, ETH3D, DIODE) differ because each protocol
prepares the data differently. Marigold's uses the native images and cameras with fixed depth ranges and crops, while
MoGe's re-renders every sample into a centered, undistorted pinhole view, cropping or zooming some sets.

\begin{center}
\includegraphics[width=\linewidth,alt={Bars per benchmark: our 32B model's AbsRel relative to MoGe-3. Ours is better on HAMMER, NYUv2, KITTI, ETH3D, iBims-1 and Sintel, and worse on DIODE.}]{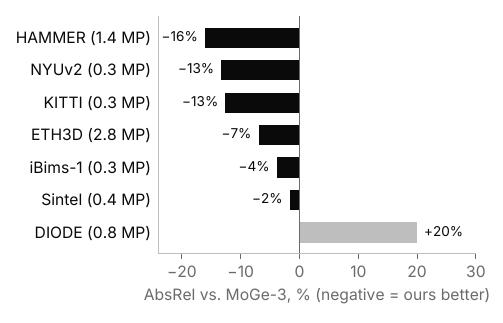}
\captionof{figure}{\modelbig{} against MoGe-3 on MoGe-3's benchmarks and alignment, per set. Black bars mark
sets where ours is better.}
\label{fig:sets}
\end{center}

\paragraph{Scale.} The same recipe was run three times with more scale each time (Fig.~\ref{fig:glance}b). A LoRA
that saw 0.2M frames scored 3.83 with 40 passes, already ahead of Marigold~V2. A full fine-tune that saw about 3.8M
unique frames, \modelsmall{}, brought it to 3.11, and moving to the 3.5$\times$ larger backbone, \modelbig{}, brought
it to 2.92 with fewer unique frames (about 3.6M). With 10 passes the same three runs score 3.99, 3.27 and 3.13
(\modelbig{} at an earlier checkpoint). The larger backbone is better on both tables, and the depth representation and
losses were the same in all three runs.

\paragraph{Cost.} Tab.~\ref{tab:protocols} sets accuracy against cost. One step of \modelbig{} takes 0.35\,s at
$512^2$ and 1.4\,s at $1024^2$ on one H200, so one prediction (40 passes) takes about 15\,s at $512^2$, and
\modelsmall{} takes 0.10\,s per step and 4.8\,s per prediction. A specialist is still much cheaper per image. MoGe-3
takes 0.41\,s at $512^2$ on the same GPU (0.06\,s of it for its 3 refinement steps), about 37 times less. We plan to
cut inference time with distillation.

\vspace{2mm}\noindent
\begin{minipage}[t]{\linewidth}\centering
\captionof{table}{Accuracy against cost. Accuracy is the mean AbsRel$\times$100 over each benchmark table (lower
is better), and time is one prediction per image at $512^2$ on one H200.}
\label{tab:protocols}
\setlength{\tabcolsep}{4pt}
\begin{tabular}{@{}lrrr@{}}
\toprule
 & \multicolumn{2}{c}{\thd{Mean AbsRel}} & \\
\cmidrule(lr){2-3}
\thd{Model} & \thd{Marigold's sets} & \thd{MoGe-3's sets} & \thd{Time}\\
\midrule
\modelbig{} & \best{2.92} & \best{3.56} & 15\,s\\
\modelsmall{} & 3.11 & 3.84 & 4.8\,s\\
MoGe-3 & 3.12 & 3.70 & 0.4\,s\\
\bottomrule
\end{tabular}
\end{minipage}

\end{multicols}
\vspace{3mm}
\noindent\begin{minipage}{\linewidth}\centering\small
\captionof{table}{MoGe-3's protocol: relative-error L1 alignment, AbsRel$\times$100 (lower is better), on seven of its nine sets (Sec.~\ref{sec:eval}). Reported: as published in MoGe-3's paper~\cite{moge3} (Table~C.1, affine-invariant depth). Our runs: run and scored by our code on the same pixels. The ten models with the lowest mean, plus the reference model as its own paper reports it. Shaded bold: best per column.}
\label{tab:moge}
\setlength{\tabcolsep}{5pt}
\begin{tabular}{@{}lrrrrrrrr@{}}
\toprule
\thd{Model} & \thd{NYUv2} & \thd{KITTI} & \thd{ETH3D} & \thd{iBims-1} & \thd{Sintel} & \thd{HAMMER} & \thd{DIODE} & \thd{Mean}\\
\midrule
\multicolumn{9}{@{}l}{\textit{Reported}}\\
Depth Pro & 3.67 & 5.12 & 4.97 & 3.23 & 15.8 & 3.30 & 4.66 & 5.82\\
InfiniDepth & 3.97 & 5.13 & 4.62 & 3.13 & 14.4 & 3.77 & 4.94 & 5.71\\
Pixel-Perfect Depth & 3.29 & 4.66 & 3.27 & 2.94 & 14.1 & 2.99 & 3.95 & 5.03\\
UniDepth V2 & 2.96 & 3.85 & 2.94 & 2.64 & 13.4 & 2.48 & 4.04 & 4.62\\
UniK3D & 2.81 & 3.81 & 2.98 & 2.54 & 13.6 & 2.47 & 3.57 & 4.54\\
MoGe-2 & 2.88 & 3.74 & 2.80 & 2.37 & 13.4 & 2.72 & 3.12 & 4.43\\
MoGe-3 (ViT-G) & 2.62 & 3.42 & 2.03 & 2.08 & 10.9 & 2.34 & \best{2.65} & 3.72\\
\midrule
\multicolumn{9}{@{}l}{\textit{Our runs}}\\
Marigold V2 & 3.34 & 4.96 & 4.83 & 2.81 & 14.88 & 2.80 & 4.18 & 5.40\\
MoGe-3 & 2.63 & 3.41 & 2.01 & 2.07 & 10.85 & 2.30 & \best{2.65} & 3.70\\
\midrule
\modelsmall{} & 2.40 & 3.53 & 2.18 & 2.20 & 11.50 & 2.22 & 2.88 & 3.84\\
\modelbig{} & \best{2.28} & \best{2.98} & \best{1.87} & \best{1.99} & \best{10.67} & \best{1.93} & 3.18 & \best{3.56}\\
\bottomrule
\end{tabular}
\end{minipage}
\vspace{5mm}

\noindent\begin{minipage}{\linewidth}\centering\small
\captionof{table}{Marigold's protocol: RANSAC alignment in log depth, AbsRel$\times$100 (lower is better). Reported: as published in Marigold~V2's paper~\cite{marigoldv2} (Table~1, one decimal; $^\dagger$ $\pi^3$ trains on ScanNet, not zero-shot). Our runs: run and scored by our code on the same pixels. The ten models with the lowest mean, plus the reference model as its own paper reports it. Shaded bold: best per column.}
\label{tab:marigold}
\setlength{\tabcolsep}{5pt}
\begin{tabular}{@{}lrrrrrr@{}}
\toprule
\thd{Model} & \thd{NYUv2} & \thd{ETH3D} & \thd{ScanNet} & \thd{KITTI} & \thd{DIODE} & \thd{Mean}\\
\midrule
\multicolumn{7}{@{}l}{\textit{Reported}}\\
InfiniDepth & 4.3 & 6.1 & 4.7 & 8.7 & 6.4 & 6.04\\
MoGe-2 & 3.1 & 5.0 & 3.1 & 6.9 & 9.6 & 5.54\\
PPD (1024) & 4.1 & 4.3 & 4.6 & 7.0 & 6.8 & 5.36\\
Lotus-2 & 3.7 & 4.1 & 4.0 & 6.7 & 6.5 & 5.00\\
FE2E & 3.8 & 3.8 & 4.3 & 6.5 & 5.6 & 4.80\\
Marigold V2 & 3.6 & 2.8 & 3.7 & 5.4 & 5.2 & 4.14\\
$\pi^3$ & 2.9 & 2.9 & 2.1$^\dagger$ & 5.9 & 4.4 & 3.64$^\dagger$\\
\midrule
\multicolumn{7}{@{}l}{\textit{Our runs}}\\
Marigold V2 & 3.58 & 2.78 & 3.65 & 5.41 & 5.07 & 4.10\\
MoGe-3 & 2.65 & 2.06 & 2.86 & \best{3.98} & \best{4.03} & 3.12\\
\midrule
\modelsmall{} & 2.07 & 2.12 & 2.56 & 4.58 & 4.22 & 3.11\\
\modelbig{} & \best{1.90} & \best{1.76} & \best{2.25} & 4.16 & 4.50 & \best{2.92}\\
\bottomrule
\end{tabular}
\end{minipage}
\vspace{5mm}

\begin{multicols}{2}

\section{Discussion}
\label{sec:discussion}
\paragraph{Toward one model.} We set out to test whether a model that only knew RGB could learn the geometry of what
it paints while remaining a generator. \modelbig{} now beats MoGe-3, a model built for this one task, at the
resolutions we trained on, and it still generates images close to those of its base model. Video and image generators
show the same pattern on other tasks~\cite{wiedemer2025,visionbanana}, which suggests that one model could learn
every modality of a scene together and keep improving with data and compute. Depth is the first modality we added,
and normals, segmentation and multi-view geometry are next.

\paragraph{Limitations.} \emph{Depth is still relative.} The prediction is defined up to a scale and shift in log
space, so \model{} gives neither metric distances nor the camera, while MoGe-3 predicts both. We plan to close this
gap by adding one more group of tokens to the same sequence that predicts the camera intrinsics. With the camera
known, the depth map becomes a point cloud from which the shift can be recovered, leaving a single global scale.
\emph{Resolution.} We train at up to about $1024^2$ pixels and do not evaluate on 2 to 8\,MP images (Spring,
Synth4K). \emph{Cost.} Each prediction takes 40 passes of a 9B or 32B transformer, about 37 times slower than MoGe-3
(Sec.~\ref{sec:results}). We accepted that trade on purpose, because the open question was whether one generative
model could match the specialists at all, and image generation has already shown how to make such models faster
through distillation. \emph{Generation drift.} Generated images move toward the look of the training data. The human
preference score of \modelbig{} falls from 26.71 to 24.30, prompt adherence drops slightly (CLIP 27.42 against 27.85)
and photorealism rises. \emph{Single images.} We evaluate one image at a time, so consistency across video frames or
views is not measured.

\balance

\end{multicols}

\clearpage
\appendix
\renewcommand{\thesubsection}{\Alph{section}.\arabic{subsection}}
\iflatexml\else\titleformat{\section}{\brock\fontsize{15.5}{18}\selectfont\color{ink}}{\Alph{section}\;·\;}{0pt}{}\fi
\section{Examples}
\label{app:qual}
All examples are from \modelbig{}, chosen from the held-out prompts and benchmark images of our evaluation.

\subsection{Image $\rightarrow$ depth}
\begin{center}
\sbox0{\includegraphics[height=\dimexpr\textheight-6.4cm\relax,width=\linewidth,keepaspectratio,alt={Rows of benchmark photographs with ground truth and predictions.}]{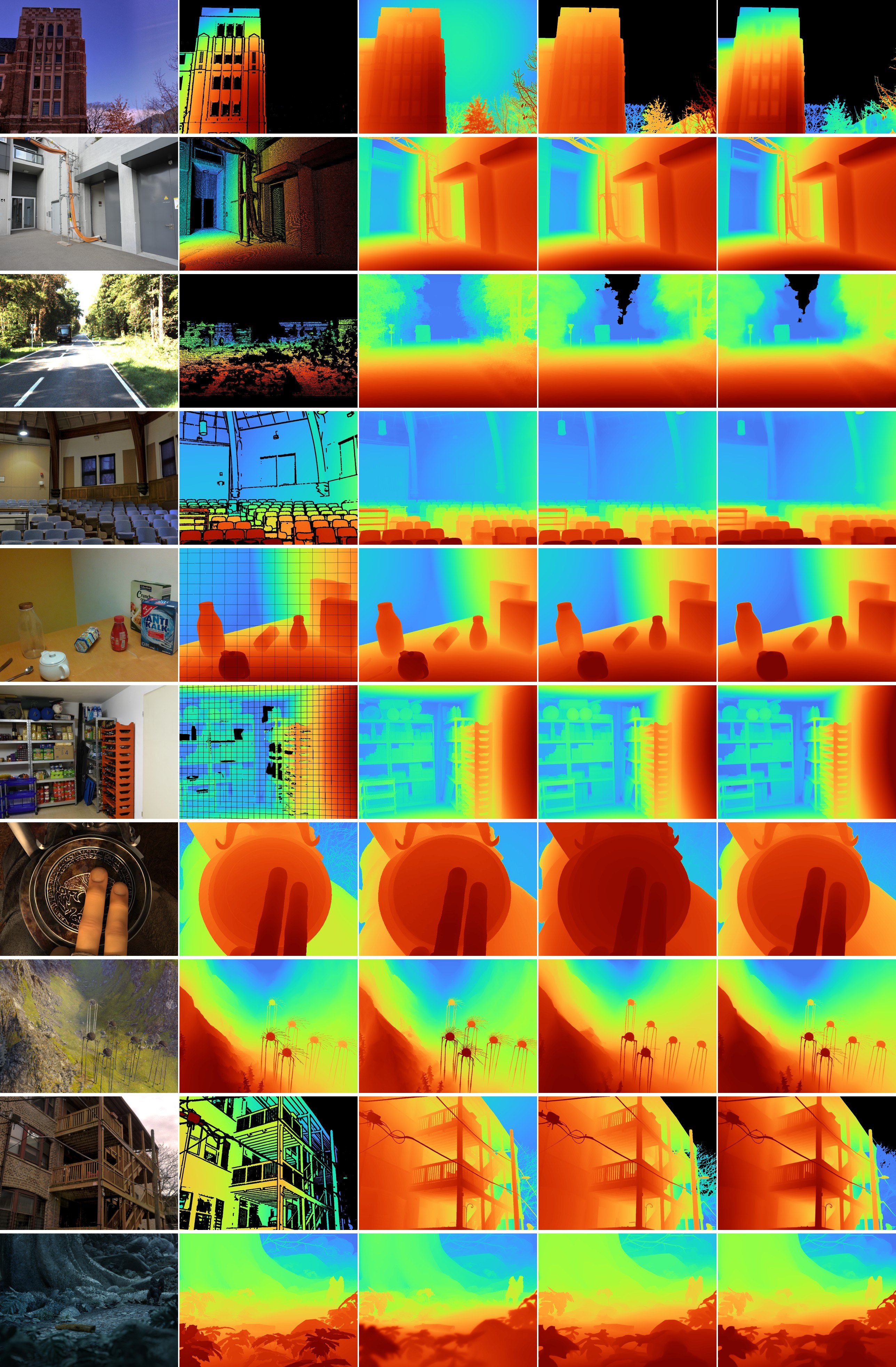}}%
\usebox0\par\vspace{1mm}
\makebox[\wd0]{\footnotesize\makebox[0.2\wd0]{Photograph}\makebox[0.2\wd0]{Ground truth}\makebox[0.2\wd0]{Marigold~V2}\makebox[0.2\wd0]{MoGe-3}\makebox[0.2\wd0]{\modelbig{}}}
\captionof{figure}{Benchmark photographs with their ground truth (black: no measurement) and each model's
prediction. Each prediction is aligned as its table scores it and colored on the ground truth's
range (near = red, far = blue). Black: no depth or sky; ours and MoGe-3 mark sky with their own sky masks, and
Marigold~V2 has no sky output.}
\label{fig:benchmarks}
\end{center}

\clearpage
\subsection{Prompt $\rightarrow$ RGBD}
\begin{center}
\includegraphics[height=\dimexpr\textheight-4.2cm\relax,width=\linewidth,keepaspectratio,alt={Generated images, each above the depth map generated with it.}]{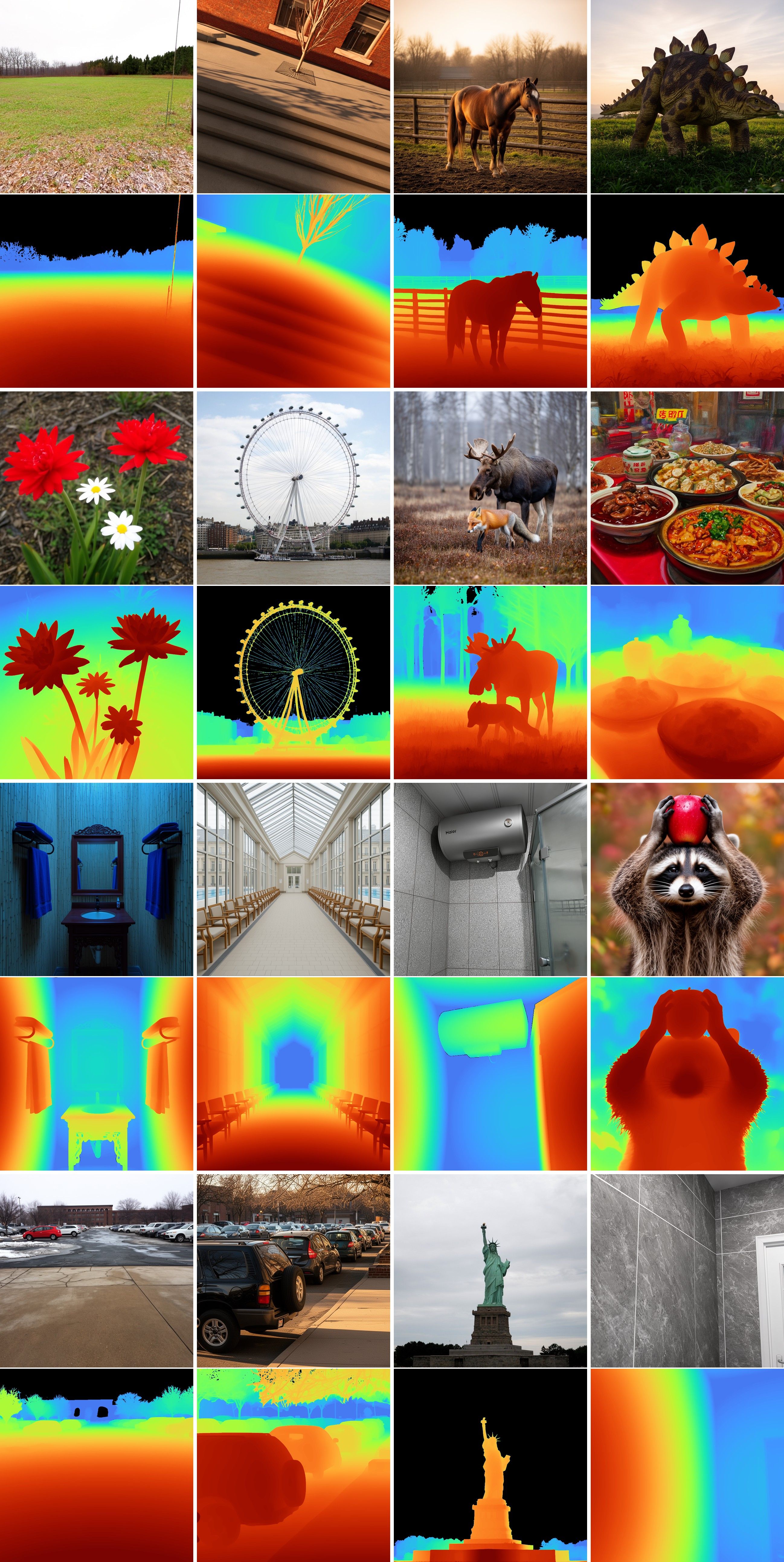}
\captionof{figure}{Prompt $\rightarrow$ image and depth, generated together from held-out test prompts; each
image above its depth.}
\label{fig:app_rgbd}
\end{center}

\clearpage
\subsection{Depth $\rightarrow$ image}
\begin{center}
\includegraphics[height=\dimexpr\textheight-4.2cm\relax,width=\linewidth,keepaspectratio,alt={Given depth maps, real photographs and painted images.}]{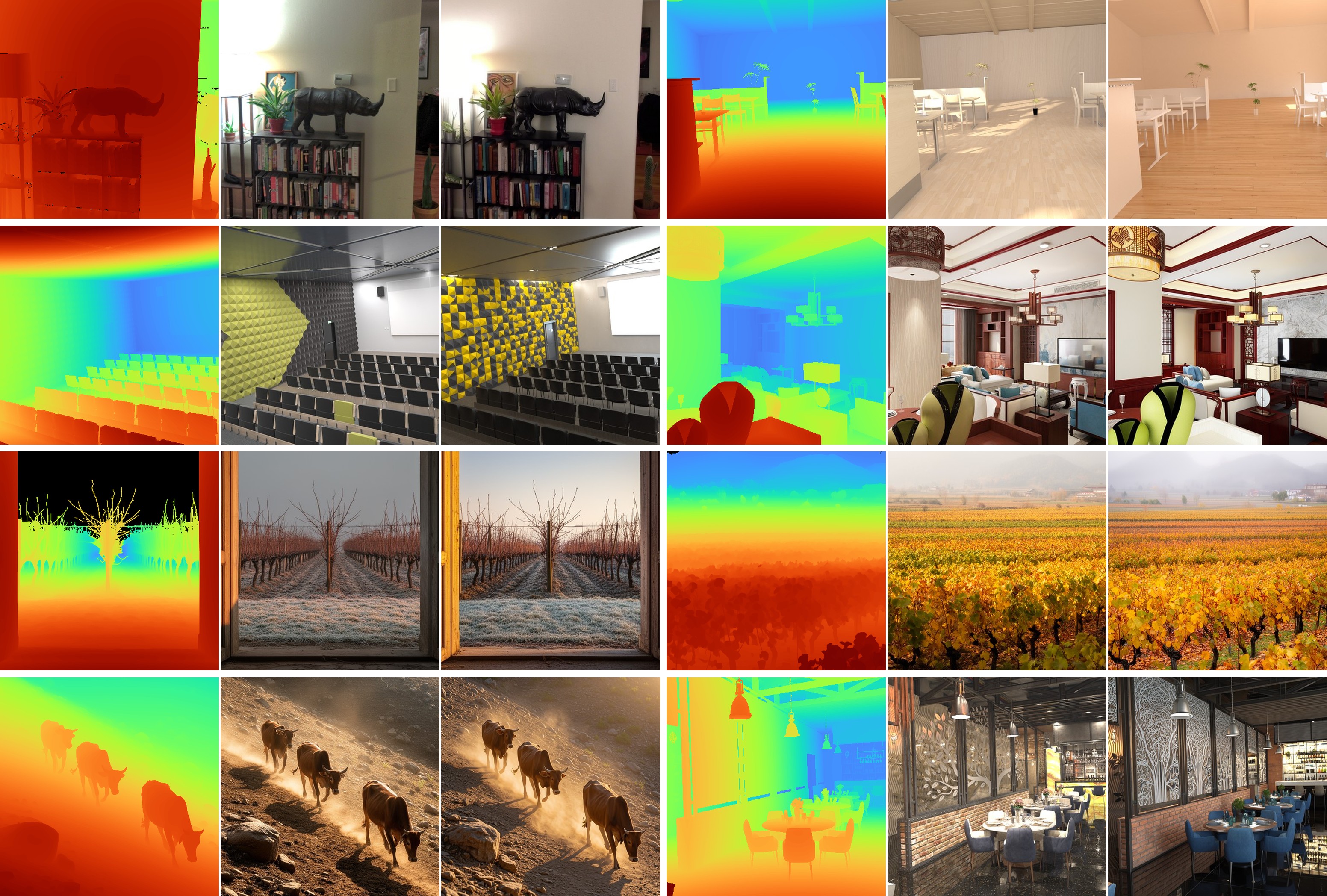}
\captionof{figure}{Depth $\rightarrow$ image on held-out frames: the given depth, the real image, and
\modelbig{}'s image painted onto the depth with the frame's caption.}
\label{fig:app_d2r}
\end{center}

\clearpage
\subsection{Base model vs.\ ours}
\begin{center}
\includegraphics[height=\dimexpr\textheight-4.2cm\relax,width=\linewidth,keepaspectratio,alt={Prompts with the untouched base model's image and ours side by side.}]{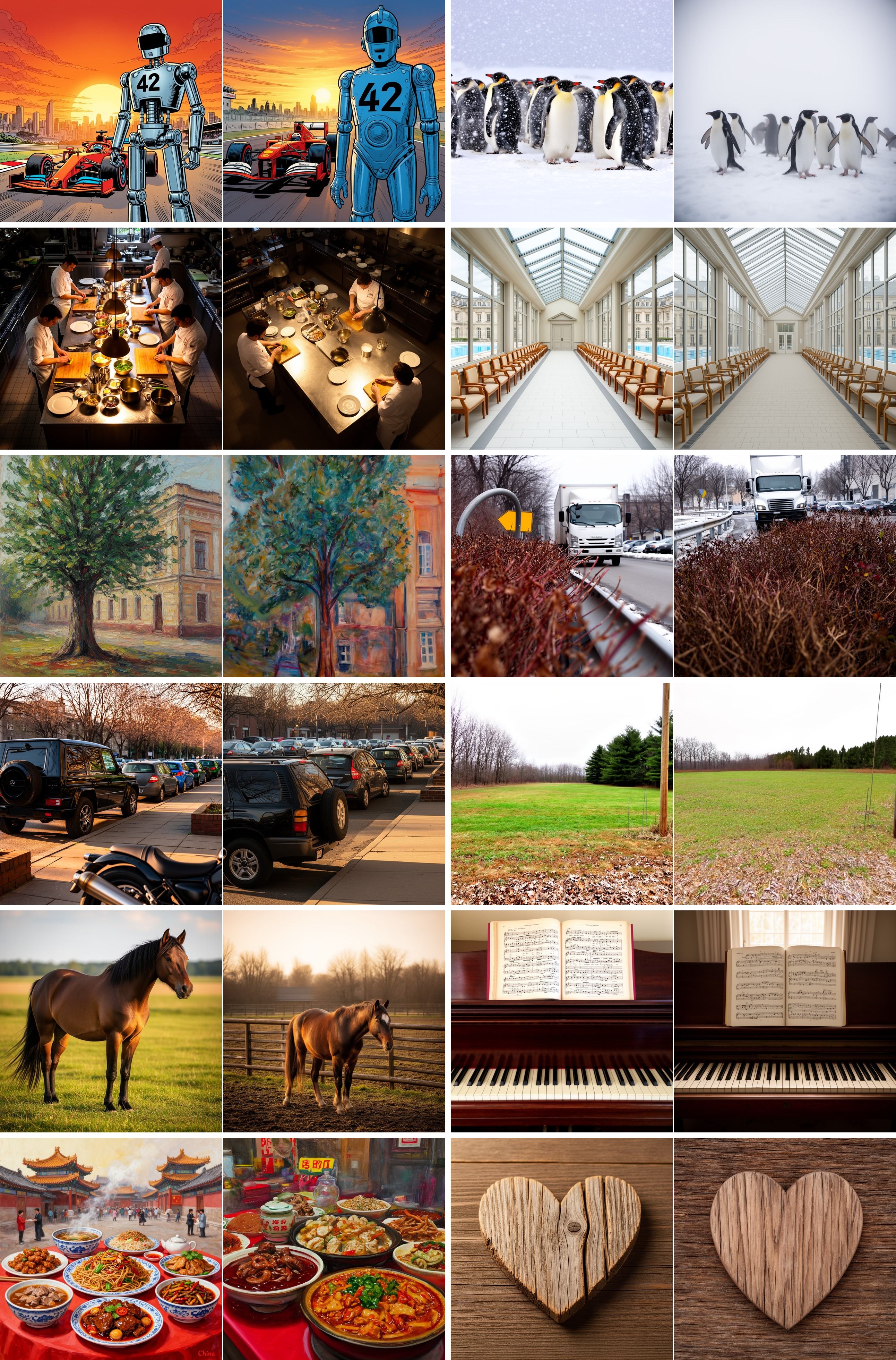}
\captionof{figure}{Same prompt and seed, left the untouched base model, right \modelbig{} (the image from a
joint RGBD generation).}
\label{fig:app_gen}
\end{center}

\clearpage
\section{Depth accuracy: $\delta_1$}
\label{app:delta1}
The same runs as Tab.~\ref{tab:moge} and Tab.~\ref{tab:marigold}, scored with $\delta_1$: the share of valid pixels
whose aligned prediction is within a factor of 1.25 of the ground truth, in \% (higher is better). Same alignments,
sets and resolutions as the AbsRel tables (Sec.~\ref{sec:eval}).

\begin{center}\small\renewcommand{\arraystretch}{0.82}
\captionof{table}{MoGe-3's protocol on seven of its nine sets, $\delta_1$ in \% (higher is better). Reported: as published in MoGe-3's paper~\cite{moge3} (Table~C.1, affine-invariant depth). Our runs: run and scored by our code on the same pixels. Same models as Tab.~\ref{tab:moge}. Shaded bold: best per column.}
\label{tab:moge_d1}
\setlength{\tabcolsep}{5pt}
\begin{tabular}{@{}lrrrrrrrr@{}}
\toprule
\thd{Model} & \thd{NYUv2} & \thd{KITTI} & \thd{ETH3D} & \thd{iBims-1} & \thd{Sintel} & \thd{HAMMER} & \thd{DIODE} & \thd{Mean}\\
\midrule
\multicolumn{9}{@{}l}{\textit{Reported}}\\
Depth Pro & 98.2 & 96.8 & 96.4 & 98.3 & 80.1 & \best{99.6} & 95.6 & 95.00\\
InfiniDepth & 97.7 & 96.2 & 96.5 & 98.5 & 81.8 & 99.1 & 95.2 & 95.00\\
Pixel-Perfect Depth & 98.4 & 97.3 & 98.6 & 98.6 & 82.0 & 99.2 & 96.6 & 95.81\\
UniDepth V2 & 98.6 & 98.1 & 98.5 & 98.4 & 83.2 & \best{99.6} & 96.5 & 96.13\\
UniK3D & 98.6 & 98.1 & 98.6 & 98.4 & 83.2 & \best{99.6} & 97.0 & 96.21\\
MoGe-2 & 98.6 & 98.1 & 99.1 & 98.9 & 82.3 & 99.5 & 97.6 & 96.30\\
MoGe-3 (ViT-G) & 98.7 & 98.4 & 99.5 & 98.9 & \best{86.9} & \best{99.6} & 98.0 & 97.14\\
\midrule
\multicolumn{9}{@{}l}{\textit{Our runs}}\\
Marigold V2 & 98.44 & 96.64 & 97.71 & 98.49 & 80.01 & 99.46 & 95.82 & 95.22\\
MoGe-3 & 98.66 & 98.40 & \best{99.54} & 98.86 & 86.86 & 99.58 & \best{98.13} & \best{97.15}\\
\midrule
\modelsmall{} & 98.76 & 98.48 & 99.13 & 98.80 & 85.30 & 99.59 & 97.59 & 96.81\\
\modelbig{} & \best{98.85} & \best{98.85} & 99.34 & \best{99.10} & 86.64 & 99.38 & 97.05 & 97.03\\
\bottomrule
\end{tabular}
\end{center}

\vspace{4mm}

\begin{center}\small\renewcommand{\arraystretch}{0.82}
\captionof{table}{Marigold's protocol, $\delta_1$ in \% (higher is better). Reported: as published in Marigold~V2's paper~\cite{marigoldv2} (Table~1, one decimal; $^\dagger$ $\pi^3$ trains on ScanNet, not zero-shot). Our runs: run and scored by our code on the same pixels. Same models as Tab.~\ref{tab:marigold}. Shaded bold: best per column.}
\label{tab:marigold_d1}
\setlength{\tabcolsep}{5pt}
\begin{tabular}{@{}lrrrrrr@{}}
\toprule
\thd{Model} & \thd{NYUv2} & \thd{ETH3D} & \thd{ScanNet} & \thd{KITTI} & \thd{DIODE} & \thd{Mean}\\
\midrule
\multicolumn{7}{@{}l}{\textit{Reported}}\\
InfiniDepth & 97.6 & 95.4 & 96.9 & 92.3 & 95.8 & 95.60\\
MoGe-2 & 98.4 & 96.2 & 98.3 & 89.9 & 84.5 & 93.46\\
PPD (1024) & 97.7 & 98.0 & 97.2 & 95.5 & 95.9 & 96.86\\
Lotus-2 & 97.6 & 98.6 & 97.2 & 94.1 & 95.5 & 96.60\\
FE2E & 97.6 & 98.7 & 97.1 & 96.0 & 96.4 & 97.16\\
Marigold V2 & 98.0 & 99.2 & 97.9 & 97.4 & 97.1 & 97.92\\
$\pi^3$ & 98.7 & 98.9 & 99.1$^\dagger$ & 96.7 & 96.8 & 98.04$^\dagger$\\
\midrule
\multicolumn{7}{@{}l}{\textit{Our runs}}\\
Marigold V2 & 98.01 & 99.15 & 97.89 & 97.42 & 97.05 & 97.90\\
MoGe-3 & 98.49 & \best{99.62} & 98.31 & 98.30 & 97.57 & 98.46\\
\midrule
\modelsmall{} & 99.25 & 99.53 & 98.71 & 98.04 & \best{98.00} & 98.71\\
\modelbig{} & \best{99.28} & 99.56 & \best{98.85} & \best{98.59} & 97.65 & \best{98.78}\\
\bottomrule
\end{tabular}
\end{center}

\end{document}